\documentclass[10pt, a4paper]{article}
\usepackage{lrec2022} 
\usepackage{multibib}
\newcites{languageresource}{Language Resources}
\usepackage{graphicx}
\usepackage{tabularx}
\usepackage{soul}
\usepackage{titlesec}
\titleformat{\section}{\normalfont\large\bfseries\center}{\thesection.}{1em}{}
\titleformat{\subsection}{\normalfont\SmallTitleFont\bfseries\raggedright}{\thesubsection.}{1em}{}
\titleformat{\subsubsection}{\normalfont\normalsize\bfseries\raggedright}{\thesubsubsection.}{1em}{}
\renewcommand\thesection{\arabic{section}}
\renewcommand\thesubsection{\thesection.\arabic{subsection}}
\renewcommand\thesubsubsection{\thesubsection.\arabic{subsubsection}}

\usepackage{epstopdf}
\usepackage[utf8]{inputenc}

\usepackage{hyperref}
\usepackage{xstring}

\usepackage{color}

\usepackage{caption}
\usepackage{subcaption}

\title{Face2Text revisited: Improved data set and baseline results}

\name{
    Marc Tanti$^1$, Shaun Abdilla$^2$, Adrian Muscat$^3$, Claudia Borg$^4$ \\
    {\bf \large Reuben A. Farrugia$^5$, Albert Gatt$^{1,6}$}
}

\address{
    $^1$University of Malta, Institute of Linguistics and Language Technology \\
    $^2$ $^4$ University of Malta, Department of Artificial Intelligence \\
    $^3$ $^5$University of Malta, Communications and Computer Engineering \\
    $^6$Utrecht University, Information and Computing Sciences \\
    \{marc.tanti, shaun.abdilla.07, adrian.muscat, claudia.borg, reuben.farrugia\}@um.edu.mt, a.gatt@uu.nl
}

\abstract{
    Current image description generation models do not transfer well to the task of describing human faces. To encourage the development of more human-focused descriptions, we developed a new data set of facial descriptions based on the CelebA image data set. We describe the properties of this data set, and present results from a face description generator trained on it, which explores the feasibility of using transfer learning from VGGFace/ResNet CNNs. Comparisons are drawn through both automated metrics and human evaluation by 76 English-speaking participants. The descriptions generated by the VGGFace-LSTM + Attention model are closest to the ground truth according to human evaluation whilst the ResNet-LSTM + Attention model obtained the highest CIDEr and CIDEr-D results (1.252 and 0.686 respectively). Together, the new data set and these experimental results provide data and baselines for future work in this area. \\
    \newline
}

\begin{document}

\maketitleabstract


\section{Introduction}

Image description generation models currently do not take into account the human element of facial description, and usually stop at either a very high-level (e.g. \textit{A blonde woman}) or give incorrect facial descriptions \cite{nezami2019facecap}. A critical part of human-generated facial descriptions is a more in-depth analysis of the facial features themselves, sometimes including inferred emotions or expressions.

Developing data specifically focusing on facial description has benefits that go beyond the image description generation task. It would potentially improve information retrieval to the extent of making it easier for more accurate facial images to be obtained when searching the web, and more importantly, it would make software and web browsing a dramatically better experience for users with visual impairment \cite{blindvisual}. It is also helpful in forensic analysis \cite{suspectface}, bridging the gap between face descriptions and what those faces actually look like. This also affects the work being done in the inverse task of generating facial images from descriptions, which would lend the power of artificial intelligence to the work currently being done by (computer-aided) sketch artists. With enough data and a powerful enough model, the subjectivity that is currently intrinsic to sketching would be balanced out, ideally resulting in a generated face which is less biased and more likely to aid with the identification of people in the area of forensics. It would also be of benefit to the arts in the reverse task - books which describe a face can automatically generate depictions of what the character should look like, depending on the textual description. Casting of actors for a film adaptation could also be aided with a similar facial generation. 

The objectives of the present work were (a) to encourage research in this direction with the development of a new data set of facial descriptions based on the CelebA data set of celebrity faces \cite{celebA}, and (b) to study the use of deep learning architectures (VGGFace/ResNet CNNs and LSTMs) for generating detailed descriptions from images of human faces. The models developed were evaluated by humans as well as using automatic metrics.

The rest of this paper is structured as follows. Section 2 provides a review of related data sets and models, mostly in the area of image description generation. Section 3 describes the development of the data set, whilst section 4 describes the baseline models. The models are evaluated and discussed in section 5, and section 6 concludes the paper.


\section{Related work}

\subsection{Image description data sets}

There is a wide variety of data sets for image description generation or image generation from descriptions. Some focus on scenes, such as MSCOCO \cite{Lin2014} and WikiScenes \cite{Wu2021}, some on fine-grained object descriptions, such as Caltech-UCSD Birds and Oxford Flowers-102 \cite{Reed2016}, and others focus on multilingual descriptions, such as Multi30k \cite{Elliott2016}.

The original Face2Text data set \cite{face2text} -- which the present work expands and improves upon -- was the first data set to focus on faces. It was based on 400 photos from the Labelled Faces in the Wild data set \cite{facesinthewild} and the descriptions were collected through crowd sourcing. Prior to this data set, the closest to a facial description data set was CelebA \cite{celebA} which is a collection of over 200k photos of celebrity faces obtained from the web, which pairs these images with data attributes such as hair colour and gender. This was followed by the Multi-Modal CelebA data set \cite{Xia2021} which consists of $30\,000$ images from CelebA together with automatically constructed descriptions from the attributes. The limitation of this data set is that, since the descriptions are artificially constructed, they do not provide `gold' annotations that give clues as to what people find salient in faces. Another facial description data set is FlickrFace11K \cite{nezami2019facecap} which consists of $11\,696$ images extracted from Flickr30K \cite{young-flickr30k}. Although the descriptions were written by humans, the images do not focus on the faces exclusively as they are scene photos and some photos contain more than one face. This made the descriptions lack the level of detail that we target in our data set.

Given the small size of the original Face2Text, the low quality face photos, and the low quality descriptions collected due to the nature of crowd sourcing, we revamped the data set to use CelebA images, and we sourced descriptions from human annotators who were hired for the purpose, and thoroughly briefed about the process.

\subsection{Image description generation models}

Image Description Generation models have the objective of generating global or dense descriptions for a given visual input, and hence require an understanding of both visual and linguistic elements. As in other areas of NLP, including vision and language processing, current image captioning models tend to be based on the pre-train-and-fine-tune paradigm, making use of Transformer-based architectures \cite{Vaswani2017} pre-trained in a task-agnostic fashion on large (usually web-sourced) data sets \cite{Sharma2018}. Examples of such models include OSCAR \cite{Li2020}, VinVL \cite{Zhang2021} and LEMON \cite{Hu2021}.

Since our goal in this paper is to establish baseline results, the remainder of this section focuses on classic attention-based encoder-decoder models, which are used in producing the baseline.

The Encoder-Decoder framework is arguably the standard model used in generating image descriptions. It works similarly to neural machine translation methods, with the image being the source and the sentence description being the target. In its most simple form, a Convolutional Neural Network (CNN) is used to encode the scene and the objects present in the image, together with their relationships. The output from the CNN is then passed into a sequence model, a Recurrent Neural Network (RNN) or derivatives of it, that acts as a conditioned language model which can be used to generate a sentence that is conditioned on the input image. For example, the Show and Tell image caption generator \cite{vinyalsshowandtell} uses a Long Short-Term Memory (LSTM) neural network to model the probability of a sentence given an input image.

Attention-based image description aims to generate suitable descriptions by paying attention only to the most visually relevant contents of an image, similarly to how primates and humans see and pay attention \cite{spratlingattention}. The first work to use attention mechanisms in image description generation was the Show, Attend and Tell image caption generator \cite{showattendandtell}, where an encoder-decoder model was fitted with an attention mechanism that would attend to salient parts of the image during the decoding process. Using an LSTM as a decoder, the attention mechanism selects visual features from the image and uses this to generate the next word in the sentence.


\section{Data collection}

At the time of publication, we have released two versions of the new Face2Text data set: version 1 and version 2. Both of these versions are publicly available\footnote{
    Data sets can be downloaded from: \url{https://github.com/mtanti/face2text-dataset}.
}. The images are not included due to copyright reasons but can be downloaded separately from the CelebA data set \cite{celebA}. The baseline facial description generator was trained on version 1.

The annotation was done in two phases, for version 1 and 2. For version 1, 4 annotators were recruited and paid at a rate of €0.14 per caption. For version 2, 11 more annotators were recruited and paid at a rate of €0.08 per caption. 

\begin{figure*}[ht]
    \centering
    \includegraphics[width=0.8\textwidth]{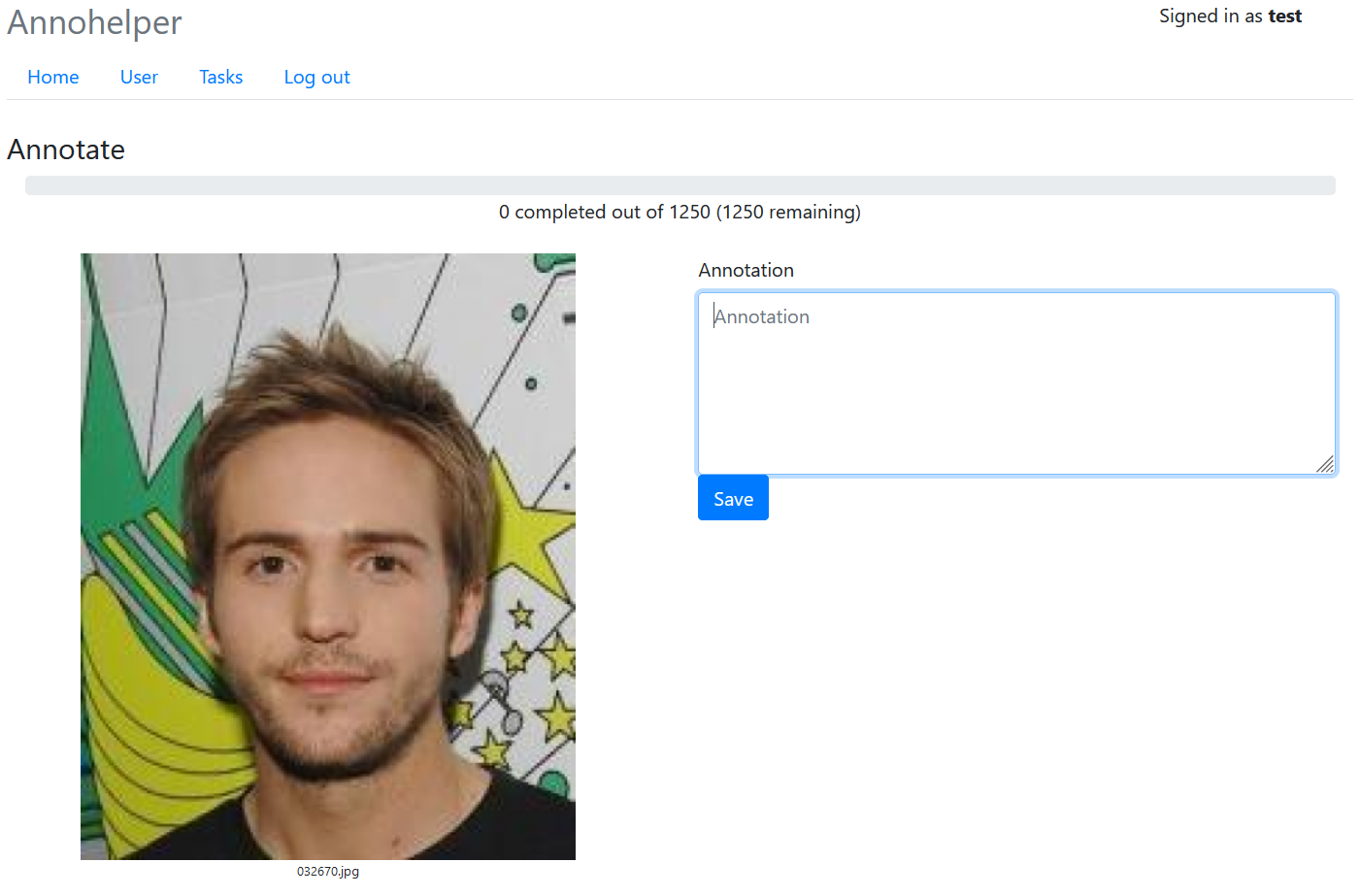}
    \caption{Screenshot of the annotation website developed for our annotators.}
    \label{figure:screenshot}
\end{figure*}

For each version, we selected a random sample of images from CelebA and stratified the sample such that the number of males and females depicted in the images was balanced. We then assigned a subset of the images to each annotator, depending on the number they were willing to annotate, such that no annotator annotated the same image more than once. The annotators then used a website, developed in-house, to write a description for each image. Annotators worked at their own pace and the data set was collected over the course of several months. Figure~\ref{figure:screenshot} shows a screenshot of the annotation tool.

The recruited annotators were students enrolled at the University of Malta. They first went through a trial run with 10 descriptions that were closely inspected before the annotators were engaged to do the entire allotment, thus ensuring quality. The instructions given to the annotators were the following:

\begin{itemize}
    \item Describe the faces as naturally as possible.
    \item Do not spend too much time thinking about what to write. Just write the description which, in your view, accurately captures the physical attributes of the face.
    \item Don't describe the background and don't make inferences about the situation of the photo or the person (such as the person's job or background).
    \item You can describe a person's facial expression or their emotional state if this is evident from the picture.
    \item Given that the images are of celebrities, do not mention the names of people you recognise.
\end{itemize}

Furthermore, the annotators were made aware that their descriptions would be made public but that the annotators' identities would not be revealed. Prior to launching the study, we obtained clearance from the University of Malta Research Ethics Committee.\footnote{\url{https://www.um.edu.mt/research/ethics/}}

\subsection{Data statistics}

\begin{figure*}[ht]
    \centering
    \begin{subfigure}[t]{0.22\textwidth}
        \centering
        \includegraphics[width=\textwidth]{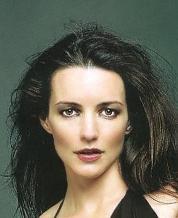}
        \caption{\textit{A woman with a chiselled jaw, prominent cheekbones, a long, narrow nose and thin eyebrows. She has long, messy, black hair and she is wearing makeup.}}
    \end{subfigure}
    \quad
    \begin{subfigure}[t]{0.22\textwidth}
        \centering
        \includegraphics[width=\textwidth]{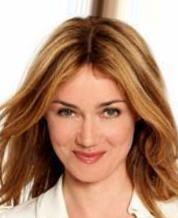}
        \caption{\textit{A woman with long amber hair with black roots, having large cheeks and a small mouth, wearing makeup and red lipstick.}}
    \end{subfigure}
    \quad
    \begin{subfigure}[t]{0.22\textwidth}
        \centering
        \includegraphics[width=\textwidth]{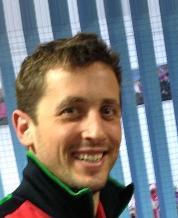}
        \caption{\textit{A man with sun-tanned face, short brown hair, big downturned eyes and a wide smile.}}
    \end{subfigure}
    \quad
    \begin{subfigure}[t]{0.22\textwidth}
     \centering
        \includegraphics[width=\textwidth]{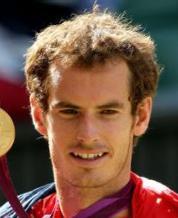}
        \caption{\textit{a white man with brown hair, open mouth and dark colored eyes}}
    \end{subfigure}

    \caption{Examples of descriptions in the data set.}
    \label{figure:description_examples}
\end{figure*}

Some examples of the descriptions obtained, together with a table of figures about the data sets are shown in Figure~\ref{figure:description_examples} and Table~\ref{table:dataset_stats} respectively. Note that version 2 of the new data set is an extension of the data in version 1. None of the data from the original Face2Text data set was used in the new data sets.

\begin{table}[ht]
    \centering
    \begin{tabular}{lrrr} 
        \hline
         & Orig. & v1 & v2 \\
        \hline
        Num. annotators & 186 & 4 & 11 \\
        Num. images & 400 & $4\,076$ & $10\,559$ \\ 
        Num. descriptions & $1\,445$ & $5\,685$ & $17\,022$ \\
        Num. tokens & $32\,619$ & $175\,555$ & $439\,291$ \\
        Num. token types & $3\,404$ & $1\,553$ & $2\,538$ \\
        Descs./image & 3.61 & 1.39 & 1.61 \\
        Descs./annotator & 7.77 & $1\,421.25$ & $1\,547.45$ \\
        Tokens/description & 22.57 & 30.88 & 25.81 \\
        Tokens/token type & 9.58 & 113.04 & 173.09 \\
        \hline
    \end{tabular}
    \caption{Quantitative summary of the Face2Text data sets. Note that `Orig.' refers to the original Face2Text data set \protect\cite{face2text} whilst `v1' and `v2' refer to version 1 and version 2 of the new data set described in this work.}
    \label{table:dataset_stats}
\end{table}


\section{Experiments}

In this section we describe the baseline face description generator models we developed using version 1 of the new Face2Text data set. As already mentioned above, the models consist of an attention mechanism using a CNN as an encoder and an LSTM as a decoder. Variations are applied to this architecture to create different models and the results are reported.

The encoder CNN is either ResNet101 \cite{heimagerecognition}, which was pre-trained on the ImageNet data set (with the task of classifying the objects in an image), or VGG-Face \cite{Schroff2015}, which was pre-trained on the VGGFace data set (with the task of face recognition). These CNNs had their dense layers at the end removed to reveal the convolution layers and extract localised visual features from the images. They were also either fine-tuned or frozen during training.

The decoder LSTM either uses attention \cite{showattendandtell} or does not. The word embeddings are either taken from GloVe \cite{Pennington2014} or are randomly initialised and fine-tuned with the rest of the model. Beam search is used to decode the sentences using beam sizes between 1 and 5.

For ease of reference, the model variants are denoted by 4-letter acronyms described in Table \ref{tab:experimentlegend}.

\begin{table}[ht]
    \begin{center}
        \begin{tabular}{r|l}
            \hline
            Character & Meaning \\
            \hline
            R & ResNet encoder \\
            V & VGG Face encoder \\
            \hline
            G & GloVe embeddings \\
            E & No Pre-trained embeddings \\
            \hline
            F & Fine-tuned encoder \\
            N & Encoder not fine-tuned \\
            \hline
            A & LSTM with attention decoder \\
            L & LSTM decoder \\
            \hline
            1-5 & Beam search size \\
            \hline
        \end{tabular}
        \caption{Character legend to the experiment variations.}
        \label{tab:experimentlegend}
    \end{center}
\end{table}


\section{Results}

\begin{figure*}[ht]
    \centering
    \includegraphics[width=0.8\linewidth]{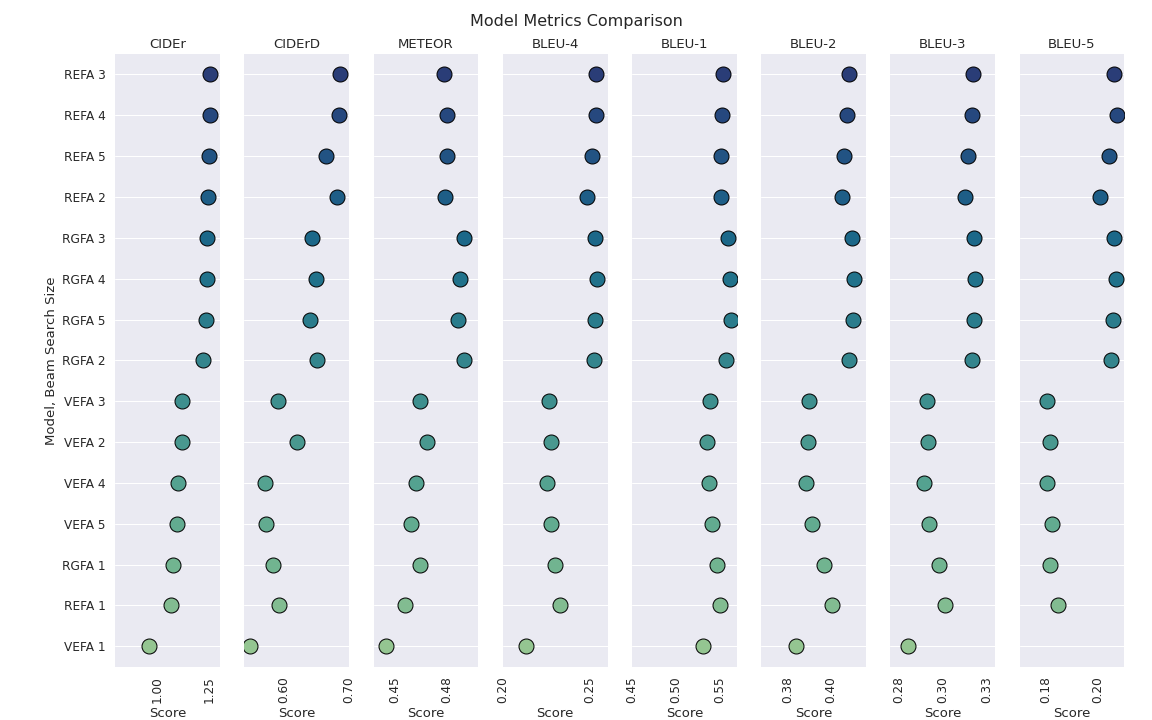}
    \caption{Swarmplot of evaluation metrics on the different variations of the face description generator.}
    \label{fig:swarmplot}
\end{figure*}

A number of evaluation metrics were applied to evaluate the performance of the face description generator. These were CIDEr, CIDEr-D, METEOR, and BLEU-1 to BLEU-5. Figure~\ref{fig:swarmplot} shows a swarmplot of the top results.

\begin{figure}[ht]
    \centering
    \includegraphics[width=0.5\linewidth]{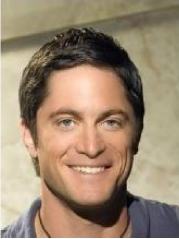}
    \caption{
        Descriptions for the best described image: \\
        Ground truth - \textit{A young man with short brown hair and blue eyes. His lips are thin and his upper teeth are visible. He is smiling} \\
        VEFA - \textit{A man with short black hair thick eyebrows a wide nose and a smile
        with dimples} \\
        RGFA/REFA - \textit{A young man with short dark hair and small dark eyes. His lips are thin and his upper teeth are visible. He is smiling}
    }
    \label{fig:best_image}
\end{figure}

The best performing model, according to CIDEr, was REFA, that is, fine-tuned ResNet CNN with randomly initialised word embeddings and attention. Further hyperparameter tuning was performed on the embedding size, LSTM size, and minibatch size of the top three variations (top three when the beam size is ignored) and the performance of the resulting models is shown in Table~\ref{table:results}. REFA, the best model after tuning, has its hyperparameters listed in Table~\ref{table:hyperparameters}. Some example descriptions of the same image, from the best-performing models, are shown in Figure~\ref{fig:best_image}.

\begin{table}[ht]
    \center
    \begin{tabular}{l|lll}
        \hline
        Model & METEOR & CIDEr & CIDEr-D \\
        \hline
        VEFA & 45.83 & 1.078 & 0.581 \\
        RGFA & \textbf{47.80} & 1.200 & 0.634 \\
        REFA & 47.06 & \textbf{1.212} & \textbf{0.662} \\
        \hline
    \end{tabular}
    \caption{Results of best three models after hyperparameter tuning using automatic evaluation.}
    \label{table:results}
\end{table}

\begin{table}[ht]
    \center
    \begin{tabular}{r|l}
        \hline
        Hyperparameter & Value \\
        \hline
        Optimiser & Adam \\
        Learning rate & $1\times 10^{-4}$ \\
        Loss function & Cross entropy \\
        Gradient clipping & 5 \\
        Batch size & 12 \\
        LSTM size & 768 \\
        Embedding size & 1024 \\
        Beam size & 3 \\
        \hline
    \end{tabular}
    \caption{Hyperparameter values of the best performing model: REFA.}
    \label{table:hyperparameters}
\end{table}

We also performed a human evaluation with 79 human evaluators. A random sample of 20 images was selected and each evaluator was asked to indicate on a 5-point Likert scale how fluent and correct (with respect to the image) each description was. Each image was accompanied by four descriptions: the generated descriptions from the top three models and the ground truth description. The highest median correctness score (equal to 4) was achieved by the RGFA descriptions (fine-tuned ResNet CNN with GloVe embeddings and attention), although these also have the highest variance. Fluency scores obtained by the RGFA were the most comparable to those obtained by the ground truth descriptions.

\begin{figure}[ht]
    \centering
    \includegraphics[width=0.5\textwidth]{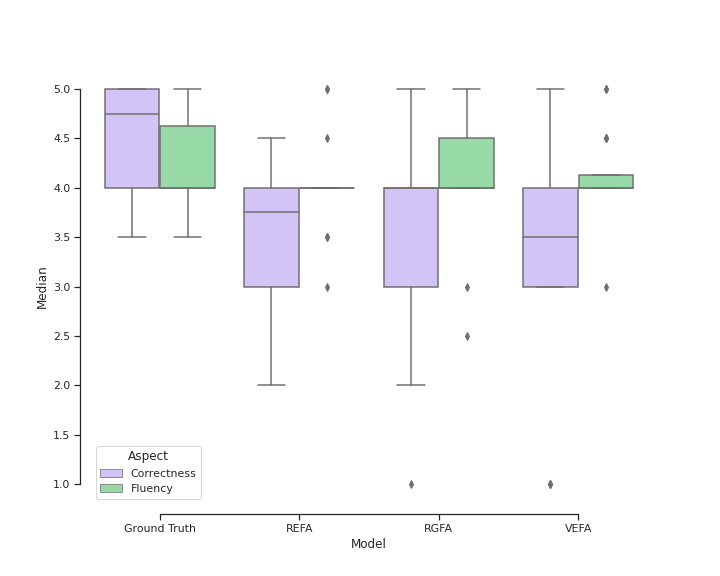}
    \caption{Results of best three models and ground truth using human evaluation.}
    \label{fig:humanground}
\end{figure}


\section{Conclusions and future work}

Our new Face2Text data set is a work-in-progress and we intend to continue adding more descriptions regularly, especially to balance the number of descriptions per image. The descriptions we have collected up to version 1 are good enough to make a strong baseline (if a pre-trained CNN is used).

We determined that, surprisingly, the ResNet CNN provides better features for a facial description generator than a face-specific CNN. Regardless of which CNN is used, it should always be fine-tuned. Whether to use pre-trained word embeddings or not does not seem to matter much but the use of attention is important. We also observe that on the face description task, one of our best performing baselines (REFA; cf Table \ref{table:results}) achieves CIDEr scores approaching those of comparable models (in the sense that they are encoder-decoder models based on recurrent units) in general scene description tasks such as MS-COCO. For example, the influential Top-Down Bottom-Up attention model with CIDEr optimisation achieves a score on MS-COCO of 1.201 \cite{anderson2018bottomup}. Future work will however need to establish baselines on more recent, Transformer-based architectures. 

In terms of further future work, the data set will benefit from more linguistic diversity, both in terms of writing style, as well as facial feature highlighting which would be useful for determining what is salient in a face.


\section{Acknowledgements}

This work was partly funded by the Malta Council for Science and Technology (MCST) - R\&I-2019-004-T, and partly by a grant from the University of Malta Research Fund. The baseline model was developed as part of Shaun Abdilla's MSc project at the University of Malta.


\section{Bibliographical References}\label{reference}

\bibliographystyle{lrec2022-bib}
\bibliography{bibliography}

\bibliographystylelanguageresource{lrec2022-bib}

\end{document}